\documentclass[conference]{IEEEtran}
\IEEEoverridecommandlockouts

\usepackage{cite}
\usepackage{amsmath,amssymb,amsfonts}
\usepackage{algorithmic}
\usepackage{graphicx}
\usepackage{textcomp}
\usepackage{xcolor}
\usepackage[hidelinks]{hyperref}
\hypersetup{colorlinks=true,citecolor=blue,linkcolor=blue,urlcolor=blue}

\graphicspath{{.}}
\def\BibTeX{{\rm B\kern-.05em{\sc i\kern-.025em b}\kern-.08em
    T\kern-.1667em\lower.7ex\hbox{E}\kern-.125emX}}
\begin{document}

\title{A Faster Approach to Spiking Deep Convolutional Neural Networks}

\author{\IEEEauthorblockN{1\textsuperscript{st} Shahriar Rezghi Shirsavar}
\IEEEauthorblockA{\textit{School of Electrical and Computer Engineering} \\
\textit{University of Tehran}\\
Tehran, Iran \\
shahriar.rezghi@ut.ac.ir}
\and
\IEEEauthorblockN{2\textsuperscript{nd} Mohammad-Reza A. Dehaqani}
\IEEEauthorblockA{\textit{School of Electrical and Computer Engineering} \\
\textit{University of Tehran}\\
Tehran, Iran \\
dehaqani@ut.ac.ir}
}

\maketitle

\begin{abstract}
Spiking neural networks (SNNs) have closer dynamics to the brain than current deep neural networks. Their low power consumption and sample efficiency make these networks interesting. Recently, several deep convolutional spiking neural networks have been proposed. These networks aim to increase biological plausibility while creating powerful tools to be applied to machine learning tasks. Here, we suggest a network structure based on previous work to improve network runtime and accuracy. Improvements to the network include reducing training iterations to only once, effectively using principal component analysis (PCA) dimension reduction, weight quantization, timed outputs for classification, and better hyperparameter tuning. Furthermore, the preprocessing step is changed to allow the processing of colored images instead of only black and white to improve accuracy. The proposed structure fractionalizes runtime and introduces an efficient approach to deep convolutional SNNs.
\end{abstract}


\section{Introduction}
\begin{figure*}[tb]
\centerline{\includegraphics[width=\textwidth]{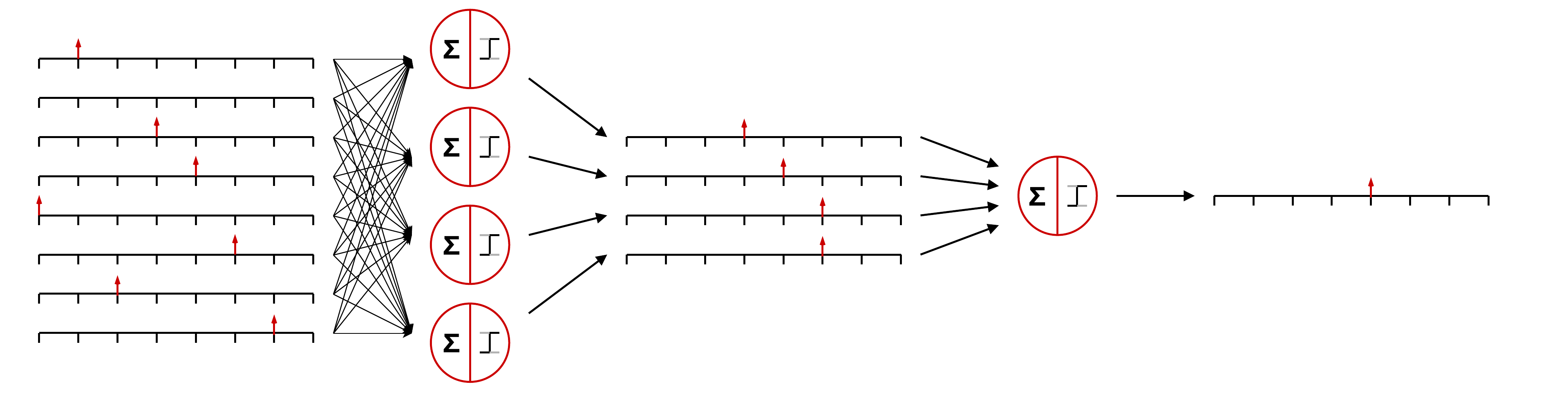}}
\caption{Demonstration of integrate-and-fire neuron with rank order coding. Here, the input is fully connected to the 4 neurons in the first layer, which are connected to a single output neuron in the output layer. There are eight time steps, and a neuron can fire at any time step or not at all. The spikes are integrated in the neurons and the internal potential increases. When this potential reaches a certain threshold, the neuron fires a spike.}
\label{fig:neuron-model}
\end{figure*}

Spiking neural networks (SNNs) are the third generation of neural networks. These networks have closer dynamics to the brain compared to the second generation of neural networks (deep neural networks or DNNs). Several electrophysiological studies emphasize the role of temporal dynamics in neural coding \cite{dehaqani_temporal_2016, dehaqani_selective_2018}. Therefore, it is critical to developing networks that contain time concepts. Because of propagating spike trains instead of continuous values, SNNs can be implemented to have low power consumption and run on embedded hardware \cite{huynh_implementing_2022}. These networks make use of the learning rules discovered in the brain \cite{markram_regulation_1997, gerstner_neuronal_1996, fremaux_neuromodulated_2016} for sample efficiency. Researchers have proposed structures and building blocks for these networks in search of ways to improve efficiency and accuracy. Neural models used to simulate biological neurons can be leaky integrate-and-fire (LIF) \cite{abbott_lapicques_1999}, spike-response model (SRM), Izhivedich, etc. Popular coding schemes for these networks are rate coding \cite{adrian_impulses_1926} and temporal coding. Different learning algorithms, from biologically plausible ones like spike-timing-dependent plasticity \cite{markram_regulation_1997, gerstner_neuronal_1996} (STDP) and reward-modulated STDP \cite{fremaux_neuromodulated_2016} (R-STDP) to algorithms with no biological roots like backpropagation and conversion from artificial neural networks to SNNS have been developed. A demonstration of the integrate-and-fire (IF) neuron combined with rank order coding \cite{thorpe_rank_1998} is shown in Fig. \ref{fig:neuron-model}.

One well-structured network that utilizes the STDP learning rule is the Kheradpisheh et al. \cite{kheradpisheh_stdp-based_2018} network. This network uses an integrate-and-fire neural model with rank order coding. This is a two-layer convolutional network each followed by a pooling layer and STDP on both layers. The input is filtered with two Difference of Gaussian (DoG) filters (one on-center and one off-center) and thresholded afterward before passing through the network. Then the input is processed by the network, and the output is generated. The output consists of the binary indication of whether the neuron has fired or not. Then a linear support vector classifier (SVM) is applied to the output. This network is able to reach 98.4\% accuracy on the MNIST dataset using unsupervised learning.

Another spiking neural network structure is proposed by Mozafari et al. \cite{mozafari_bio-inspired_2019}. The neural model and coding are the same as in the previous network, but the learning algorithm differs in the third layer. This network has three convolutional layers and uses rank order coding. The first two layers of this network are trained with STDP, and the third layer is trained with the R-STDP learning rule. This network can achieve 97.2\% accuracy using its own native classifier that makes decisions based on the category that the neuron with the highest potential belongs.

Despite the progress made in this field, the developed networks are not yet comparable in terms of accuracy and speed. The models need to be biologically plausible, have a dynamic that is able to solve complex machine learning tasks, and have computational efficiency in order to compete with current deep learning networks. This work is a biologically plausible network integrated with machine learning methods that tries to reduce the accuracy gap while improving the runtime speed of SNNs. It takes a significant step towards creating SNN models that are able to solve state-of-the-art machine learning tasks with more robustness and lower energy consumption.

\section{Methods}
The proposed network structure is based on previous works introduced in the previous section. Some modifications are made to improve the runtime of the network while improving the accuracy. In this section, the building blocks for the network are introduced. Then, the proposed network structure is explained. Finally, the evaluation methods of the network are discussed.

\subsection{Building Blocks}
The LIF model is a parallel combination of a resistor and a capacitator. Initially, the neuron will be in the resting state and have $u_{rest}$ potential. The neuron potential will rise when it receives input from synaptically connected neurons and emit a spike when the potential reaches the threshold $V_{th}$. Then, its potential is reset to the resting potential. Integrate-and-fire neuron model is used in the proposed network. The IF model can be formulated as below:
\begin{equation}
V_i(t) = V_i(t-1) + \sum_j^O I_j(t-1) W_{ij}
\end{equation}
Where $V_i$ is the internal potential of $i$th output neuron, $O$ is the number of output neurons, $I$ is the binary input to the network indicating spikes, and $W$ is the synaptic weight between the output neuron $i$ and the input neuron $j$. After the potentials are calculated, if an output neuron passes a threshold $V_{th}$, it fires.

In rate coding, the information is transmitted as the rate of firing of the neurons. The window of time needed to transmit information might be lengthy in this coding, and there is less sparsity of spikes. However, temporal coding transmits information through the timings of the spikes in a small window of time and with better sparsity compared to rate coding. Rank order coding is a temporal coding scheme that uses relative timings of the spikes. In rank order coding, only their relative timing to each other is considered. This scheme has a lighter computational complexity.

STDP learning rule adjusts the connections between neurons and causes learning. This algorithm utilizes the temporal differences of pre- and post-synaptic neurons to strengthen or weaken the synaptic connections. In this algorithm, when a pre-synaptic neuron fires before (after) the post-synaptic neuron, the strength of their connection increases (decreases). The rule can be formulated as:
\begin{equation}
\Delta W_{ij} =
\begin{cases}
A^{+}_k \times (W_{ij} - L_k) \times (U_k - W_{ij}), & t_j \leq t_i \\
A^{-}_k \times (W_{ij} - L_k) \times (U_k - W_{ij}), & t_j > t_i \\
\end{cases}
\end{equation}
Here, $A^{+}_k$, $A^{-}_k$, $U_k$, and $L_K$ are the positive and negative learning rates and lower and upper bounds of the stabilization process, respectively. Afterwards, the weight updating occurs this way:
\begin{equation}
W^+_{i,j} = max(L_k, min(U_k, \Delta W_{ij}))
\end{equation}

\subsection{Proposed Structure}
\begin{figure*}[tb]
\centerline{\includegraphics[width=\textwidth]{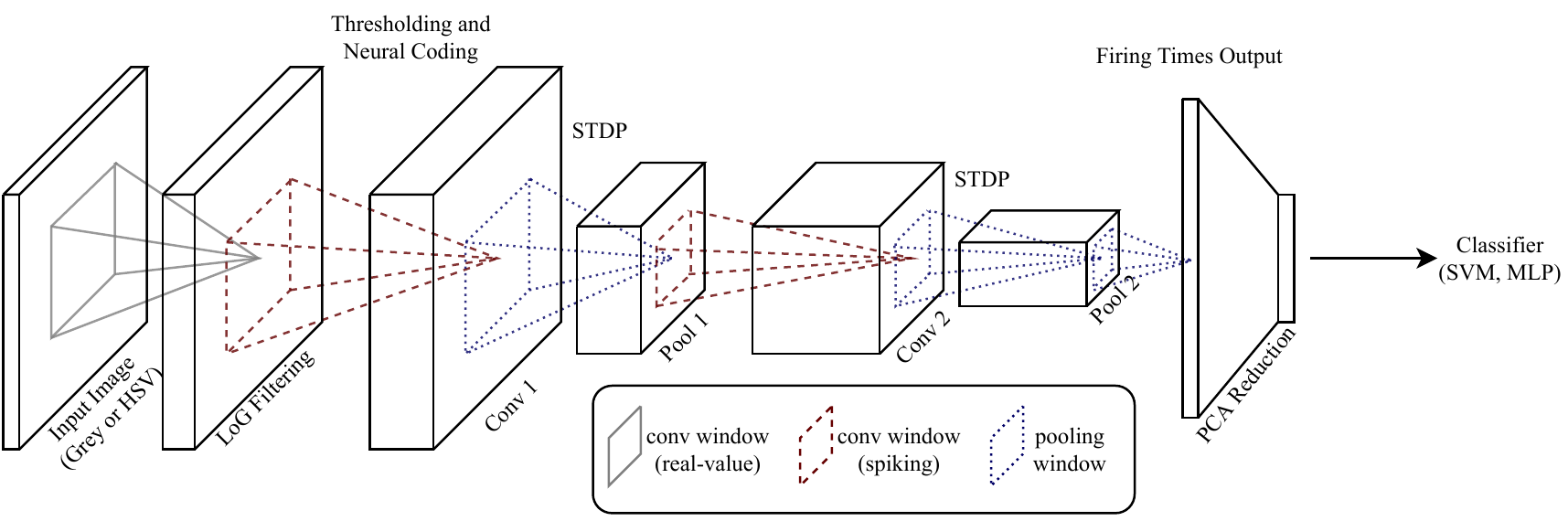}}
\caption{The overall structure of the network. The input is filtered with the LoG filter and thresholded. Then the input is coded with rank order coding. The network consists of two convolutional layers each followed by a pooling layer. Then the network output is passed through a dimension reduction operation, and the output is finally given to the classification layer.}
\label{fig:network-diagram}
\end{figure*}

Fig. \ref{fig:network-diagram} shows the structure of the network. The network input is filtered by 3 Laplacian of Gaussian filters with standard deviations of 0.471, 1.099, and 2.042. Then, values below 0.01 are reset to zero to improve the accuracy. The input values are coded into 15time steps to embed the time concept using rank order coding. The proposed network consists of two convolutional layers each followed by a pooling operation. The first convolution layer has a window size of 5, a stride of 1, and a padding of 2, and the second convolution layer has a window size of 3, a stride of 1, and a padding of 1. The window size of the first and the second pooling layers are 2 and 3 (stride is the same as window size), respectively. The synaptic weights of the convolution layers are initialized using $N\sim(0.5, 0.02)$ distribution. The threshold values of the integrate-and-fire functions are found using a grid search algorithm. Then, an SVM classifier is used to classify the outputs of the network after training.

The first change made to the structure of the network is related to the output of the network. The proposed network uses the firing times of the neurons as network output. Then the number of time steps is subtracted from the timing values so that higher values correspond to earlier firing time and more excitation. This output improves accuracy compared to the binary output indicating that a spike is fired or not.

The following change includes using a dimension reduction algorithm to speed up the classification process. Since the network outputs are sparse, principal component analysis (PCA) can be used to reduce the number of dimensions of the network output. This allows the SVM classifier to run faster and with no accuracy drop when using non-linear SVM and a small amount when using linear SVM.

Another change made to this network is an improved training process. Controlling the learning rates and their increment and quantizing the weights helps reduce the number of iterations on the dataset (only one iteration on the MNIST dataset, meaning that every layer sees every sample only once). This significantly reduces the training time and may improve classification accuracy. The quantization process involves weights above 0.5 being set to 1 and 0 otherwise.

Finally, the processing of colored images is improved. Previous networks only process black and white images, but the color values in real-world images contain valuable information. The images are first converted to an HSV code model to isolate color, contrast, and light values to take advantage of the image color information. Then the LoG filter is applied separately to each channel of the HSV input. In each category, 5 object instances are randomly selected for the training phase and the rest are only seen during the testing.

\subsection{Evaluation Methods}

The simulations are implemented using the Spyker \footnote{\url{https://github.com/ShahriarSS/Spyker}} library. The system used to perform the simulations on has Intel Core i7-9700K with 64G memory and Nvidia Geforce GTX 1080 Ti with 12G memory (Ubuntu 18.04). The evaluation of the network is done using two datasets. 30 samples for the runtime and accuracy of the networks are recorded and compared using statistical tests. In order to check for significance in the results, we test whether the two-sample mean difference confidence interval (99.9\%) contains zero. The confidence intervals are calculated using standard deviation.

\begin{figure}[htbp]
\centerline{\includegraphics[width=\columnwidth]{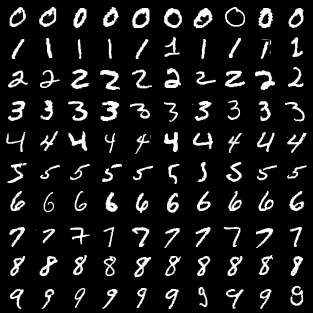}}
\caption{Sample images from the MNIST dataset. It has 10 categories for each digit. Randomly selected images from the training set are shown in the figure. It is clear to see that each row corresponds to a category.}
\label{fig:mnist-sample}
\end{figure}

The first dataset is the MNIST consisting of images of handwritten digits. It contains 60000 training and 10000 predetermined testing images with 28$\times$28 size. Sample images from the dataset are shown in Fig. \ref{fig:mnist-sample}. For this dataset, three variants of the network are proposed that differ in the number of channels. The first one is a small network with 50 and 100 channels in the convolutional layers, respectively. The second network is a medium network with 100 and 200 channels, and the third is a large network with 200 and 400 channels. The networks are trained in one iteration, and the positive and negative learning rates are 0.0004 and -0.0003, respectively, and are doubled every 2000 samples until the positive learning rate reaches 0.15.

\begin{figure}[htbp]
\centerline{\includegraphics[width=\columnwidth]{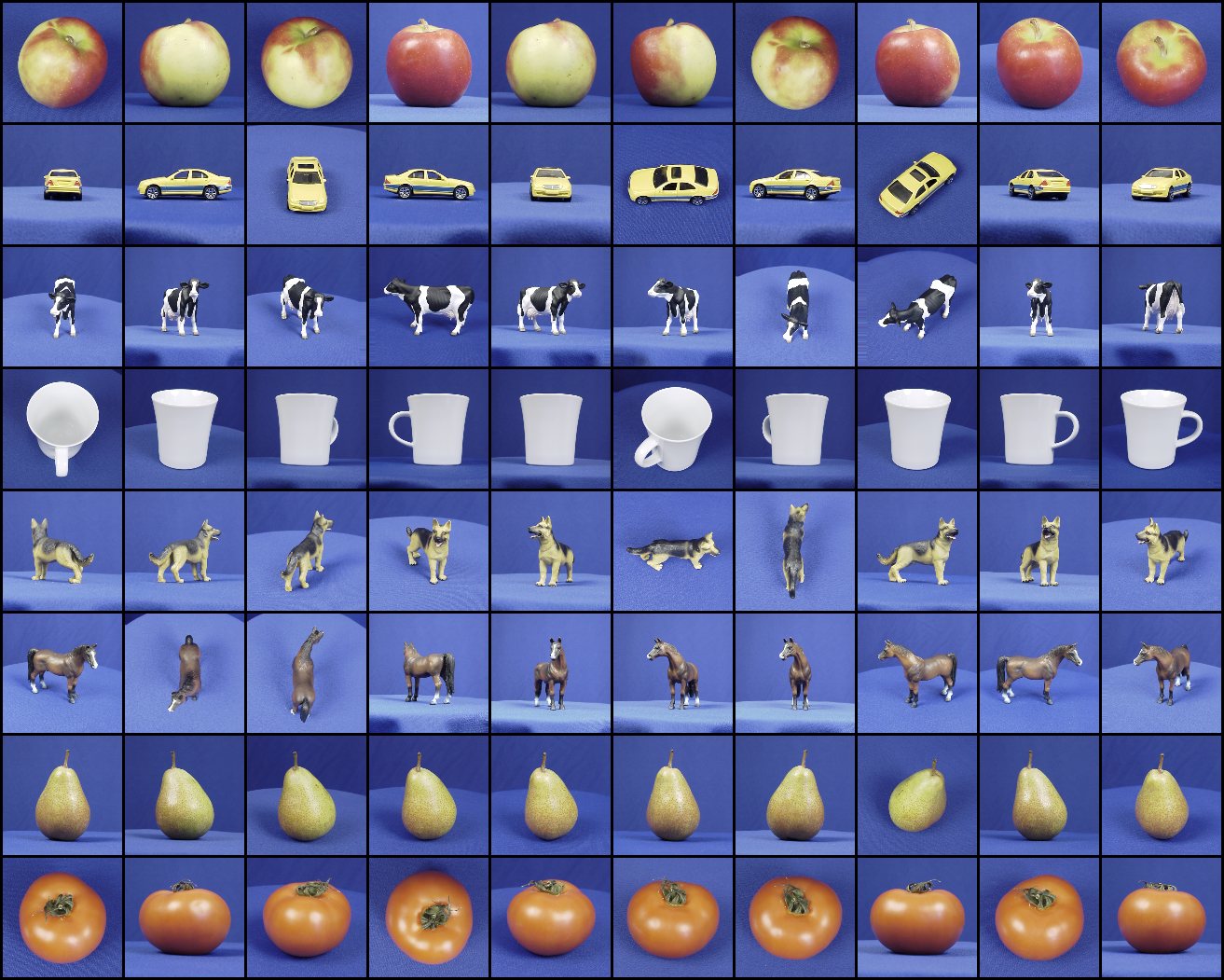}}
\caption{Sample images from the ETH-80 dataset. It has 8 categories of objects, each containing 10 real-world objects. Sample images from different angles of the first object in each category are displayed in each row.}
\label{fig:eth80-sample}
\end{figure}

ETH-80 is the second dataset containing images of different viewpoints from 8 categories of objects. Each category includes 10 object instances and 41 images of different viewpoints. The dataset is split into two equally sized parts for training and testing. The images are scaled down to 64$\times$64. Sample images from the dataset are shown in Fig. \ref{fig:eth80-sample}. Small changes to the network are made to make it suitable for this dataset. These changes include filtering the input with 9 LoG filters with standard deviations of 0.45, 0.5, 0.55, 0.95, 1, 1.05, 1.95, 2, and 2.05, setting the cutoff threshold to 0.0025, and increasing the number of channels of the convolutional layers. The network is trained for 5 iterations, and learning rates are 0.005 and -0.005 and doubled every 410 samples until the positive rate reaches 0.1. It has been suggested \cite{falez_improving_2020} that using zero-phase component analysis (ZCA) Whitening can improve performance on real-world images. This transformation is also tested replacing the LoG filter.

The network is compared to the ResNet-50 \cite{he_deep_2016} model on the ETH-80 dataset. The ResNet is tested with pretrained weight and a non-linear SVM classifier instead of the final layer. Then the ResNet is fine-tuned with L2 regularization \cite{ng_feature_2004} and a dropout layer before the final classifier layer to avoid overfitting. Training is done for 20 iterations, and the maximum test accuracy during the training is recorded. The learning rate and regularization strength are found using a grid search algorithm.

\section{Results}
\subsection{Network Training}
\begin{figure}[htbp]
\centerline{\includegraphics[width=\columnwidth]{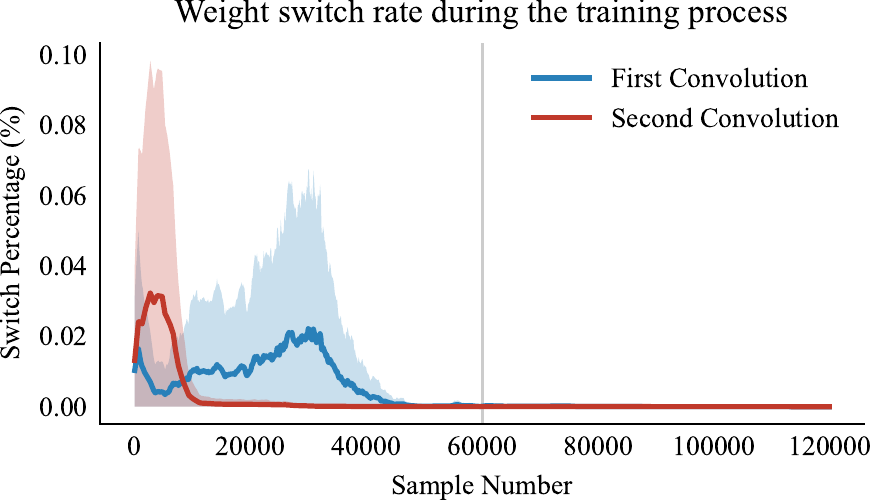}}
\caption{Plot of the average (of 30 samples) weight switch rate during the training process on the MNIST dataset. The plots are for the two convolutional layers and are passed through a moving average filter with a window size of 11. The network is trained two times on the MNIST dataset and has seen 120000 samples in total. The grey line separates the first and the second iteration on the Dataset. Error bars plotted are 95\% confidence intervals.}
\label{fig:switch-plot}
\end{figure}

The training process of the network is explained in the methods section, and here we will suggest a stopping criteria. As was explained, each layer of the network is trained, and its weights are quantized afterwards. One important aspect of training is knowing when to stop. The quantizing process introduces a great criterion. The initial weights are scattered around 0.5, and as the training progresses, they converge to values below or above this initial value. Then the weights are set to zero if they are below 0.5 and one otherwise. The proposed criterion is to watch the frequency of the switching of the weights above and below 0.5 and stopping when the switching almost stops. This is effective as the weight updating process changes the final result only when switching happens. For example, Fig. \ref{fig:switch-plot} shows the average weight switching rate for the layers of the proposed network on the MNIST dataset. As it can be seen, the second iteration on the dataset does not contribute significantly. This is why the network is only trained once on the MNIST dataset.

\subsection{MNIST Dataset}
\begin{figure}[htbp]
\centerline{\includegraphics[width=\columnwidth]{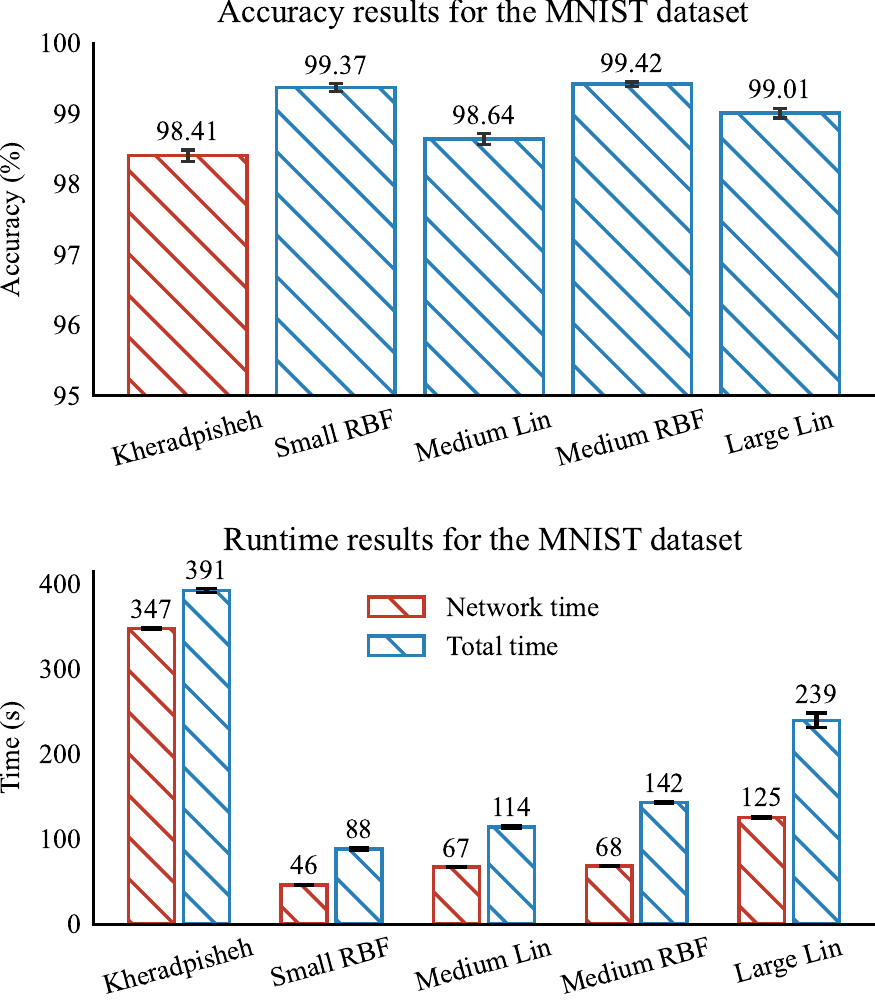}}
\caption{Comparison plots of the networks on the MNIST dataset. Here, the Kheradpisheh et al. network with a linear classifier is compared to variations of the proposed network. Classifiers used for the network are linear SVM and SVM with RBF kernel. The error bars are standard deviations.}
\label{fig:network-mnist}
\end{figure}

The comparisons for the MNIST dataset can be seen in Fig. \ref{fig:network-mnist}. The confidence interval test outputs the interval [-0.649, 0.181] for the difference of means in accuracy and [268.345, 286.655] for the difference of means in total runtime between the Kheradpisheh et al. network and the proposed network. These results show that the accuracy difference is not significant (there is no accuracy drop), and the runtime of the proposed network is significantly faster (3.43 times faster on average). The same comparisons for the large network give [-0.96, -0.25] for accuracy and [121.366, 183.234] for total runtime, indicating that the large network is significantly faster and better performing. Also, the results for the non-linear classifier are 99.372$\pm$0.055 (SD) for the small network and 99.421$\pm$0.033 (SD) for the medium network. These results show that the network is capable of achieving high accuracies combined with powerful classifiers.

\subsection{ETH-80 Dataset}
\begin{figure}[htbp]
\centerline{\includegraphics[width=\columnwidth]{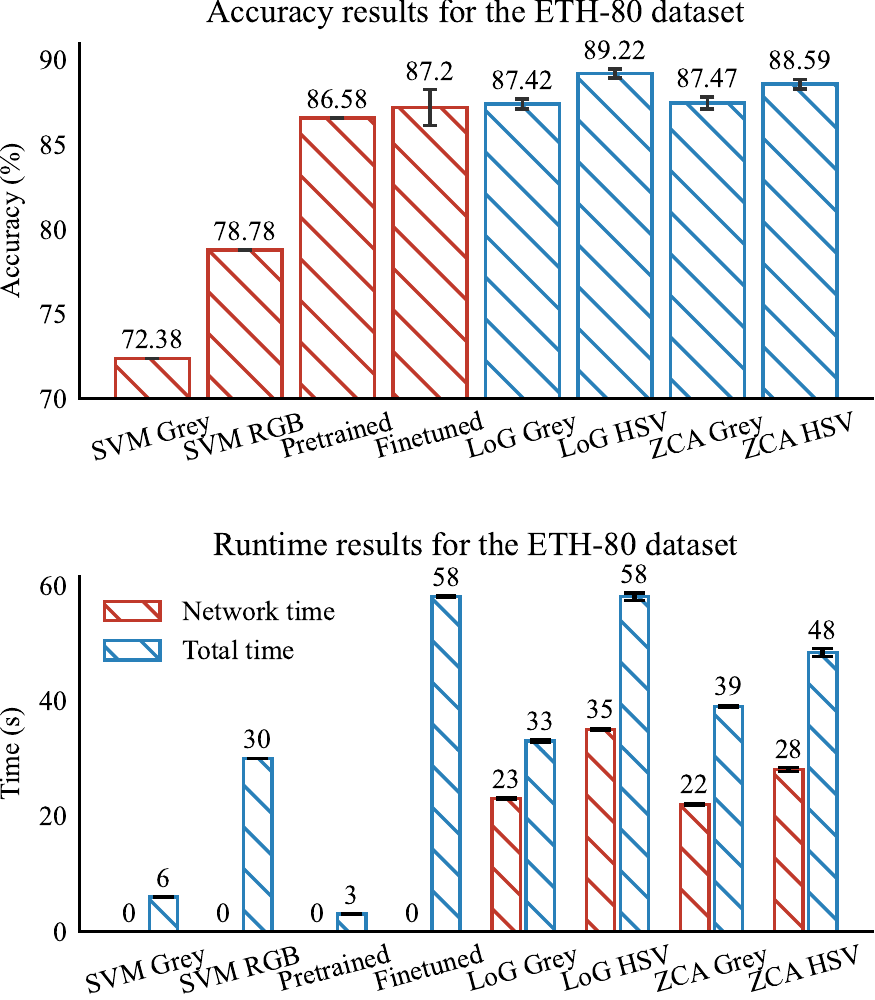}}
\caption{Comparison plots of the networks on the ETH-80 dataset. Here, the ResNet-50 network is compared to variations of the proposed network. Variations include greyscale and HSV models of the input image and LoG filter and ZCA Whitening operation. SVM classifier is used with RBF kernel. The error bars are standard deviations.}
\label{fig:network-eth}
\end{figure}

The results for the ETH-80 dataset are shown in Fig. \ref{fig:network-eth}. Only using SVM to classify this dataset results in 72.378 for greyscale images and 78.780 for RGB images provided for a baseline. The interval of the test for comparing the greyscale network and the HSV network is [-3.282, -0.327] indicating significant accuracy improvement. The result for testing LoG against ZCA Whitening is [-0.841, 2.108] for accuracy and [7.447, 11.886] for total runtime using the HSV model. This indicates that there is no accuracy drop when using ZCA Whitening and runtime is faster due to the number of channels being smaller after transformation (2 against 54 for ZCA Whitening and LoG transformations respectively). The interval for LoG with HSV network is [88.277, 90.166] and does not include 86.585 which indicates that the proposed network performs better than pretrained ResNet. Comparing the same network with fine-tuned ResNet gives [-1.83, 5.874] for accuracy and [-1.665, 1.665] for total runtime which indicates no significant difference in accuracy and runtime.

\section{Discussion}
In this paper, we introduced a spiking neural network architecture. The SNN is based on previous spiking neural networks and uses integrate-and-fire neurons, rank order coding, and the STDP learning rule. We explained the modifications made for improvements and their effects. The network is compared to the previous work on the MNIST dataset. The results indicate that the network is able to achieve higher accuracy while reducing the runtime. The networks are also evaluated on the ETH-80 dataset, and the results indicate that the network is comparable to the ResNet on this dataset.

The signals processed by the network are binary spike trains. The quantization process causes the weight to be binary. Together, these create a binary network. Having a binary network that performs well is quite desirable in different applications. Since the weights of the network are binary, they can be efficiently stored on the memory (ideally one bit per parameter). Furthermore, using small integer types to represent both signals and weight creates the advantage of lower bandwidth usage and results in runtime speedup. Moreover, optimized routines for binary multiplication and such can be used to have a faster network. These optimizations create excellent opportunities for fast and energy-efficient implementation on neuromorphic hardware.

The aim of the previous works was to create a biologically plausible network that performs well in machine learning tasks. This work takes this goal and brings it closer to reality. As a network that learns to extract features in an unsupervised manner, the proposed network performs well using the STDP learning rule. Despite mentioned features of the network, it stands to be further improved. This work uses tools from machine learning, however, the PCA transformation and non-linear SVM classifier do not have biologically plausible roots. This work uses a trade-off between biological modeling and performance in machine learning tasks. Further work might be able to find biological replacements for the mentioned tools.

Biological plausibility is an important aspect of SNNs. Having a plausible network opens up new possibilities in the world of neuroscience. More specifically, the functionality of the brain can be described and explored by experimenting with the dynamics and the structure of the proposed network. In addition, the time dynamics of the network can provide possible explanations of how temporal coding, in general, and rank order coding, in specific, can convey information and what their strengths and their limitations are.

SNNs are not yet comparable to DNNs, but this work takes a significant step toward closing the gap between the two. Future work can include exploring more complicated neural models, coding schemes, and learning rules to find a structure that shows the true potential of spiking neural networks.

\end{document}